\documentclass{bmvc2k}
\usepackage{amssymb}   
\usepackage{pifont}    
\usepackage{booktabs}

\title{Open-Vocabulary Semantic Segmentation in Remote Sensing via Hierarchical Attention Masking and Model Composition}

\addauthor{Mohammadreza Heidarianbaei}{
heidarianbaei@ipi.uni-hannover.de}{1}
\addauthor{Mareike Dorozynski}{dorozynski@ipi.uni-hannover.de}{1}
\addauthor{Hubert Kanyamahanga}
{kanyamahanga@ipi.uni-hannover.de}{1}
\addauthor{Max Mehltretter}
{mehltretter@ipi.uni-hannover.de}{1}
\addauthor{Franz Rottensteiner}{
rottensteiner@ipi.uni-hannover.de}{1}

\addinstitution{
 Institute of Photogrammetry and GeoInformation\\
 Leibniz University Hannover\\
 Germany
}

\runninghead{ Heidarianbaei, ET AL.}{Open-Vocabulary Semantic Segmentation}

\begin{document}

\maketitle

\begin{abstract}
In this paper, we propose ReSeg-CLIP, a new training-free Open-Vocabulary Semantic Segmentation method for remote sensing data. 
To compensate for the problems of vision language models, such as CLIP in semantic segmentation caused by inappropriate interactions within the self-attention layers, we introduce a hierarchical  
scheme utilizing masks generated by SAM to constrain the interactions at multiple scales.  
We also present a model composition approach that averages the parameters of multiple RS-specific CLIP variants, taking advantage of  a new weighting scheme that evaluates representational quality using varying text prompts. 
Our method achieves state-of-the-art results across three RS benchmarks without additional training. \newline \url{https://github.com/aemrhb/ReSeg-CLIP}.
\end{abstract}

\section{Introduction}
\label{Introduction}
\begin{figure}[t]
    \centering
    \includegraphics[width=.65\textwidth]{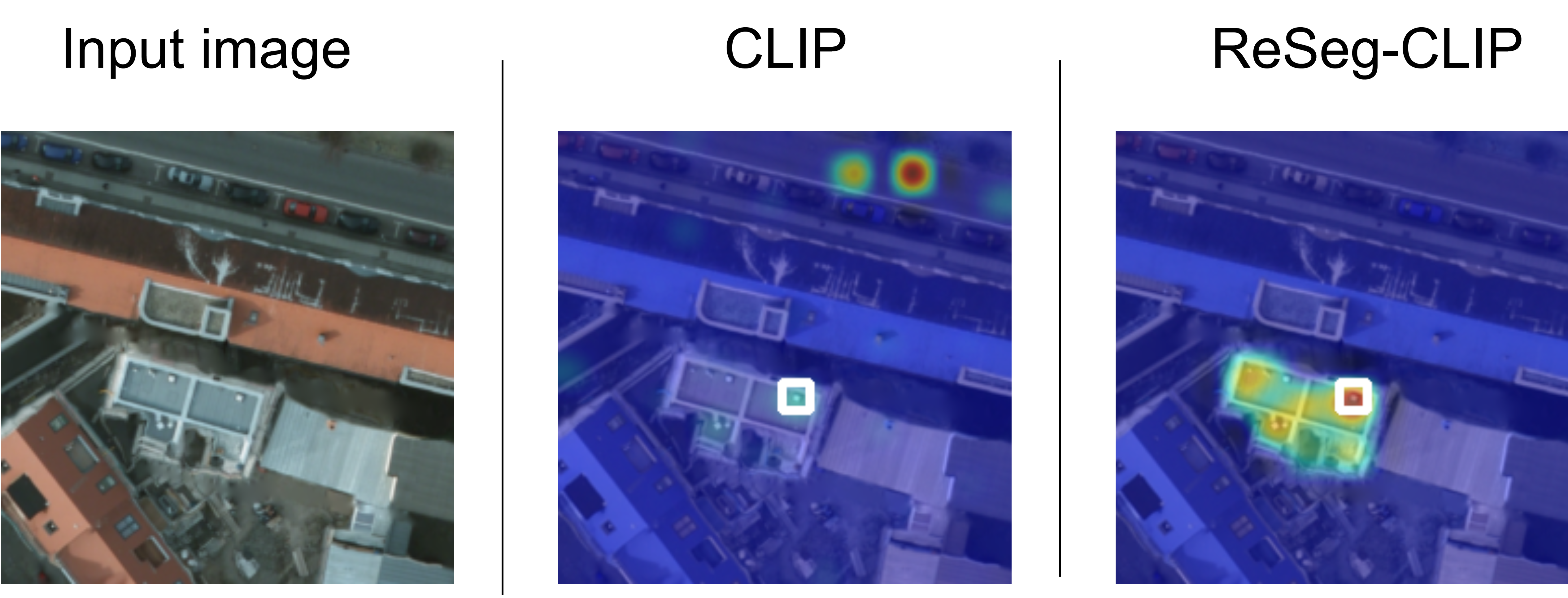}
    \caption{Example of distorted inter-patch attention for a selected patch (white squares). Left to right:  input image,  attention map obtained by the original CLIP vision encoder \cite{radford2021learning}, and the attention map obtained by our method. Blue corresponds to low attention,  red to high attention in relation to the selected patch. CLIP often assigns high attention to arbitrary patches without any relevance for the selected patch. Applying our method results in high attention concentrated on patches associated with the same object as the selected patch.}
    \label{fig:atten}
\end{figure}
Semantic segmentation is the task of assigning a class label to each pixel in an image, e.g., representing land cover in the context of remote sensing (RS). 
Despite recent advancements  \cite{lv2023deep, nedevschi2024semantic}, existing methods face two fundamental challenges: they typically require large sets of training data to perform well,   
and models trained on a specific dataset often do not generalize well to other domains. 
Recently, vision language models (VLMs) such as CLIP \cite{radford2021learning} and ALIGN \cite{jia2021scaling} have emerged as promising tools to overcome these limitations. 
Trained via contrastive learning to align images and text in a shared embedding space, these models exhibit strong zero-shot performance in image classification. 
This has motivated adaptations for Open-Vocabulary Semantic Segmentation (OVSS), resulting in models that can recognize categories beyond those seen during training.
Following the adaptation of CLIP for OVSS, early research  focused on fine-tuning 
to enhance pixel-level predictions \cite{yu2023convolutions, ghiasi2022scaling}. 
However, fine-tuning is often limited to smaller and less diverse datasets, typically leading to a reduction in the model's zero-shot capacity \cite{yu2023convolutions}.  
In response, training-free approaches have been proposed,  
mainly focusing on natural images \cite{bousselham2024grounding,shao2024explore,wang2024sclip,li2025closer,zhang2024corrclip, hajimiri2025pay}. 
Training-based adaptations of VLMs for  OVSS in RS are proposed in \cite{cao2024open, dutta2025aeroseg,chen2023toward};   Li et al.  \cite{li2024segearth}, though not training CLIP itself,  
 introduces an upsampling module that requires training. 
Observing this lack of completely training-free OVSS solutions in RS, we propose {\bf ReSeg-CLIP,} a new such method for high-resolution RS imagery which is based on two main contributions:  
\begin{itemize}
    \item A \textbf{hierarchical masking strategy} to refine attention computations by imposing constraints based on hierarchical segmentation results obtained by the Segment Anything Model (SAM) \cite{kirillov2023segment}, addressing problems of existing methods to obtain accurate pixel-wise predictions. Introducing these masks at different vision encoder stages, context is considered at different scales while mitigating the impact of unrelated patches.      
    \item A \textbf{combination of multiple domain-adapted CLIP variants} for OVSS, to improve the generalization capabilities of existing models.  For that purpose, we propose the \textbf{Prompt Variant Separation Margin}, a new metric quantifying each model’s semantic representational quality by exploiting synthetic text prompts, and use it to compute model-specific weights for the averaging of the model parameters.
\end{itemize}

Refinement is needed because VLMs like CLIP align text with global image features (via the \texttt{[CLS]} token in ViTs \cite{dosovitskiy2021an}), causing attention weights to overlook semantically related regions \cite{li2025closer}.
For instance, Figure~\ref{fig:atten} shows some patches ("outlier patches" \cite{shao2024explore}; white squares)  that attract disproportionately high attention from the rest of the image when using CLIP;  
focusing attention on such irrelevant regions causes problems for dense prediction. 
Consequently, several studies \cite{bousselham2024grounding,shao2024explore,wang2024sclip,li2025closer}  have tried to refine attention scores such that semantically related patches attend more strongly to one another. 
Although these methods boost inter-patch correlations, they still suffer from patches interacting with unrelated regions \cite{zhang2024corrclip}.
Zhang et al. \cite{zhang2024corrclip} constrain attention to regions defined by SAM masks but only at a single scale. To address varying object sizes, we extend their approach with a hierarchical masking strategy that enables the model to capture information across multiple levels. 

Our second contribution mentioned above is relevant because  CLIP, pretrained on natural images, often underperforms on RS data due to a significant domain gap. 
Prior work \cite{liu2024remoteclip1, wang2024skyscript, zhang2024rs5m}  tried to solve this problem by fine-tuning CLIP on RS data. 
However, our own experiments based on GeoRSCLIP \cite{zhang2024rs5m} and RemoteCLIP \cite{liu2024remoteclip1} show that these models still struggle to generalize across classes unseen during training, an essential requirement for OVSS. 
Inspired by   model composition techniques such as \cite{wortsman2022robust, chronopoulou2302adaptersoup, zhang2023composing, kozal2024continual,daheim2024model}, we thus propose 
to combine several CLIP models, 
each fine-tuned on a different RS dataset, by averaging the model parameters and applying the combined model for inference. 
We introduce a new metric called Prompt Variant Separation Margin (PVSM), and we use this metric for computing the weights to be used for averaging. 
PVSM measures 
the representational quality of the individual models based on the variability of the text embeddings generated for different text prompts related to the same class. 

\section{Related work}

\textbf{Vision Language Models} seek to learn joint representations from both visual and textual data, thereby enabling cross-modal understanding and reasoning. CLIP \cite{radford2021learning} and ALIGN \cite{jia2021scaling} accomplish this by contrastive learning. 
Early work on applying VLMs to RS data fine-tuned CLIP 
using RS5M, a RS-specific dataset \cite{zhang2024rs5m}. 
Subsequent works tried to obtain better models by fine-tuning them on more curated RS datasets \cite{liu2024remoteclip1, wang2024skyscript}. 
Nevertheless, VLMs continue to perform poorly on semantic segmentation. In RS, this challenge is compounded by the limited scale and diversity of datasets, restricting zero-shot performance \cite{liu2024remoteclip1}.

\vspace{1ex}\noindent\textbf{Open-vocabulary semantic segmentation} aims to assign a class label to every pixel in an image, specifying the set of classes  by textual descriptions at test time. 
Existing VLM-based OVSS methods fall into three categories:
(1) Training-based adaptations \cite{Li2022LanguagedrivenSS, dong2023maskclip, luo2023segclip, xu2022groupvit}, which often generalize poorly due to limited datasets. 
(2) Two-stage approaches, which first generate mask proposals and subsequently apply VLMs \cite{ding2022decoupling, xu2022simple}, are limited by pretraining on full images. 
(3) Training-free methods that modify the computation of self-attention. 
Li et al. \cite{li2025closer} reveal inconsistencies in the relations between semantic regions formed by self-attention layers, 
with variants such as GEM \cite{bousselham2024grounding}, ClearCLIP \cite{lan2024clearclip} 
and SCliP \cite{wang2024sclip}, 
aiming to enhance the attention mechanism. 
Instead of modifying the attention mechanism, potentially introducing discrepancies between training and inference, CorrCLIP \cite{zhang2024corrclip} uses SAM to restrict the spatial extent of patch interactions (though only at a single scale). 
All these works are primarily developed for natural images and do not account for the unique characteristics of RS data.

\vspace{1ex}\noindent\textbf{Open-vocabulary semantic segmentation for RS}
is addressed in \cite{zermatten2025learning, zhang2024segclip}, via contrastive training for pixel-text alignment and the combination of text embeddings from the CLIP text encoder with features from an image encoder, respectively.
However, such methods do not leverage existing VLMs trained on large text-image datasets.  
This is achieved in \cite{ye2025towards}, where  CLIP is combined with a specialist RS image branch in a dual-stream image encoder. 
Cao et al. \cite{10962188} use CLIP to generate orientation-adaptive similarity maps, followed by some refinement layers. 
Dutta et al. \cite{dutta2025aeroseg} incorporate  visual features from SAM to enhance semantic representations. 
However, all of these methods require training to achieve a good performance. 
To the best of our knowledge, SegEarth-OV \cite{li2024segearth} is the only approach for OVSS in RS which is claimed to be training-free; 
however, while its CLIP-based predictions need no training, its upsampling module still does. In contrast, our method is entirely training-free.

\vspace{1ex}\noindent\textbf{Model merging} aims to combine independently trained models of identical architecture, 
often employing a linear interpolation between the model parameters 
\cite{ruppert1988efficient,izmailov2018averaging,zhang2019lookahead, wortsman2022robust}.  
Ilharco et al. \cite{ilharco2023editing} define task vectors as the differences between fine-tuned and pre-trained weights to capture task-specific directions and to enable model control via simple arithmetic operations; Zhang et al. \cite{zhang2024knowledge} extend this by learning the weights for computing a linear combination. 
Recent studies propose compositional approaches for low-rank adaptation \cite{huang2024lorahub, wu2024mixture}, as well as sample-wise interpolation strategies \cite{lu2024twin, cheng2025dam}. 
Typically introducing learnable parameters, these methods are not training-free; also, they neither address OVSS nor RS. 
In contrast, we compose an RS-adapted model merging for OVSS in a training-free manner. 

\vspace{1ex}\noindent\textbf{Discussion:} Most related to our work are \cite{li2024segearth} and \cite{zhang2024corrclip}. While in \cite{zhang2024corrclip}, SAM masks are used to limit the patch interactions at a single scale; we use SAM masks at multiple scales to consider different context regions. 
Li et al. \cite{li2024segearth} also address OVSS in RS, but focus on improving the spatial resolution of a prediction obtained by a CLIP variant, requiring training an upsampling module.  We improve the CLIP-based predictions without any training.

\section{ReSeg-CLIP}
\label{Method}

\begin{figure*}[t]
    \centering
    \includegraphics[width=1\textwidth]{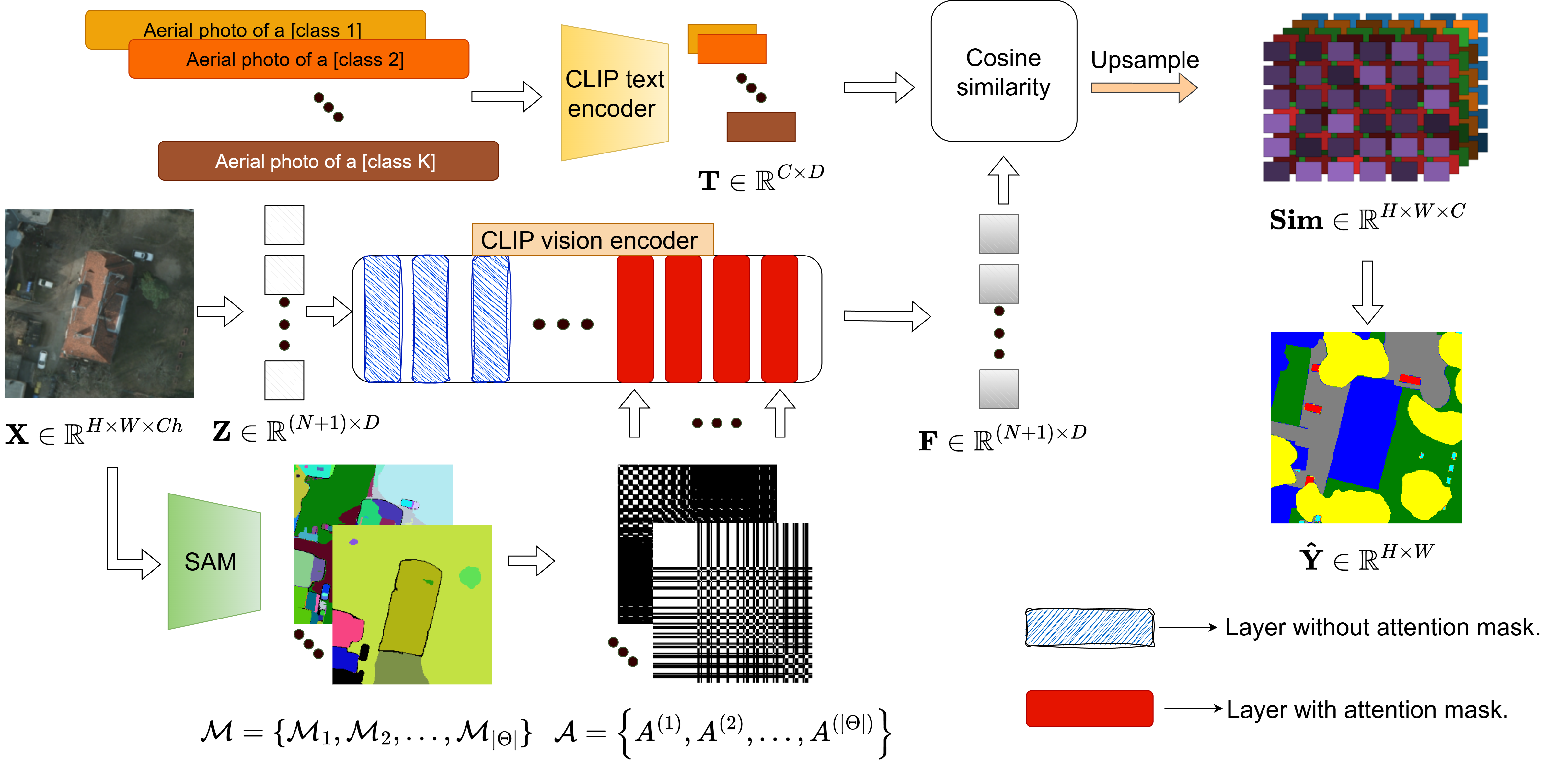}
    \vspace{-1.8em}
    \caption{
    ReSeg-CLIP consists of CLIP-based vision and text encoders and SAM. 
    The input image $\mathbf{X}$ is processed by both the vision encoder and SAM, resulting in features $\mathbf{F}$. SAM produces hierarchical masks $\mathcal{M}$, converted into attention masks $\mathcal{A}$ for the final vision encoder layers (red blocks). Text prompts for each class are encoded into embeddings $\mathbf{T}$, which are compared with the features $\mathbf{F}$ via cosine similarity. The results are upsampled to score map $\textbf{Sim}$, and the  segmentation $\hat{\mathbf{Y}}$ assigns to each pixel the class with highest similarity.
}
    \label{fig:widefig}
\end{figure*}

The goal of our proposed OVSS method, ReSeg-CLIP, is to map an RS input image $\mathbf{X} \in \mathbb{R}^{H \times W \times 3}$ to a dense label map $\mathbf{\hat{Y}} \in \mathbb{R}^{H \times W}$ (see Fig.~\ref{fig:widefig}) in a training-free manner, i.e., we rely on pre-trained CLIP models and do not introduce training at any other stage. 
$H$ and $W$ denote the image height and width, respectively, and the number of channels is 3, because CLIP is pre-trained on RGB images. 
Besides proposing, to the best of our knowledge, the first entirely training-free OVSS method for RS, our main contributions concern (1) a hierarchical guidance of attention in the image encoder, aiming to achieve interactions between semantically related patches (Sec. \ref{Refining}), and (2) a new model merging approach, aiming to enhance the model's generalization capabilities (Sec. \ref{PVSM}).
 
Before presenting $\mathbf{X}$ to the image encoder, a ViT composed of $L$ blocks, it is 
partitioned into  $N= \frac{H \cdot W}{P^2}$ disjoint patches \(\{\mathbf{x}_i\}_{i=1}^{N}\), $\mathbf{x}_i \in \mathbb{R}^{P \times P \times 3}$ 
that are flattened and projected 
to a sequence \(\mathbf{Z} = [\text{CLS}, \mathbf{z}_1, \mathbf{z}_2, \dots, \mathbf{z}_N] \in \mathbb{R}^{(N+1) \times D}\) of patch embeddings $\mathbf{z}_i  \in \mathbb{R}^D$ ($D$: embedding dimension) and a class token $\text{CLS}$. 
$\mathbf{Z}$  is processed by the modified vision encoder (see Sec. \ref{Refining} for details),  
resulting in image embeddings $\mathbf{f}_i\in\mathbb{R}^D$, where the set of all $\mathbf{f}_i$ is denoted by $\mathbf{F} \in \mathbb{R}^{(N+1) \times {D}}$.
The text encoder, a standard Transformer~\cite{vaswani2017attention}, receives class-specific base prompts.
Here, a base prompt is defined to be a 
text string (e.g., \textit{"an aerial image of [$c$] in the city"}) for all semantic classes \( c \in \{1, \dots, C\} \) (\( C \): total number of classes).
This results in text embeddings $\mathbf{t}_c \in \mathbb{R}^D$, where  
the set of all $\mathbf{t}_c$ is denoted by $\mathbf{T} = [\mathbf{t}_1, \mathbf{t}_2, \dots, \mathbf{t}_C] \in \mathbb{R}^{C \times D}$.
In contrast to the standard CLIP, relying on the $\text{CLS}$ token, the patch level predictions are obtained by computing the cosine similarity between image embeddings $\mathbf{F}$ and text embeddings $\mathbf{T}$ via  
\begin{equation}
    \text{Sim}_{i,c} = \frac{\langle \mathbf{f}_i \mathbf{t}_c\rangle}{\|\mathbf{f}_i\|_2 \, \|\mathbf{t}_c\|_2} \in\mathbb{R},
\end{equation}
resulting in the similarity map $\textbf{Sim}_{\text{low}}\in \mathbb{R}^{H/P \times W/P \times C}$. 
$\textbf{Sim}_{\text{low}}$ is bilinearly upsampled
to the original image resolution, resulting in  
$\textbf{Sim}\in \mathbb{R}^{H \times W \times C}$, and the pixel predictions are obtained by $\hat{Y}_{(x,y)} = \arg\max_c \text{Sim}_{(x,y),c}$, where $\text{Sim}_{(x,y),c}\in \textbf{Sim}$ is the score for class $c$ at pixel $(x,y)$.

\subsection{Refining the attention map of CLIP}
\label{Refining}

The main goal of the proposed refinement of CLIP's attention maps for OVSS is 
to enhance feature interactions among patches from relevant regions and suppress interference with irrelevant ones.  
To do so, we 
use class-agnostic masks provided by SAM to generate attention masks that constrain feature aggregation in the CLIP vision encoder. 
For generalization across scenes, containing both long-range and fine-grained patterns, 
we propose a hierarchical masking strategy to enable multi-scale feature aggregation, by varying the SAM mask generator and producing coarse masks at earlier stages in the encoder to encourage broad attention, and fine-grained masks at later stages to emphasize detailed semantic structures. 

We restrict the attention of the last \(r = 1, 2, \dots, |\Theta|\) layers (cf.~Fig.\,\ref{fig:widefig}), by utilizing SAM masks with different hyperparameter configurations \(\theta_r \in \Theta\) (\(\Theta\): set of all considered configurations; cf.~App.\,A.1). 
The  \(L_u= L - |\Theta| \) initial layers of the vision encoder are not modified. 
$|\Theta|$ is a hyperparameter, enabling control over the depth at which attention constraints are introduced. 
Each mask generator segments an input image $\mathbf{X}$ into $Q_r$ regions, encoded by a set of binary masks $\mathcal{M}_{\theta_r} =\{M_{1_r},M_{2_r},\dots,M_{Q_r}\}$, where each mask $M_{q_r}\in\{0,1\}^{H\times W}$ indicates a distinct region.
These masks are combined to form a label image $\mathbf{S}_r \in \{0, 1, \ldots, Q_r\}^{H \times W}$, where each pixel $(x, y)$ is labeled with its corresponding region index $RI_r$: $S_r(x, y) = q_r$ if $M_{q_r}(x, y) = 1$, and $S_r(x, y) = 0$ if the pixel does not belong to any segment (i.e., background).

The dominant RI of each patch $\mathbf{x}_i\subset \mathbf{X}$ is obtained via majority voting. Let \( \mathbf{RI_r} \in \mathbb{R}^{H/P \times W/P } \) denote the patch-level RI,
\( a, b \in \{0, 1, \dots, N\} \) the indices over the input sequence $\mathbf{Z}$, where index 0 corresponds to the class token and indices \( 1 \) to \( N \) to patch tokens. The mask ${\mathbf{A}^{(r)} \in \{0,1\}^{(N+1) \times (N+1)}}$, applied to the attention mechanism of the vision encoder at layer $l = L_u + r$, prevents high attention between patches not being in the same regions:
\begin{equation}
    \mathbf{A}_{(a,b)}^{(r)} = 
    \begin{cases}
    1 & \text{if } \left( a = b = 0\right) \lor \left[ \left(a > 0\right) \land \left(b > 0\right) \land  \left(\mathbf{RI}_{r_a} = \mathbf{RI}_{r_b} \right)\right]\\
    0 & \text{otherwise}
    \end{cases}\,,
\end{equation}
where the class token is restricted to attend only to itself (case \( a = b = 0 \)).
The process is repeated independently for each set of hyperparameters \( \theta_r \in \Theta \), each yielding a segmentation-aware attention mask. 
The final set of attention masks for the last $|\Theta|$ layers with restricted attention 
is represented as $\mathcal{A} = \{ \mathbf{A}^{(1)}, \mathbf{A}^{(2)}, \dots, \mathbf{A}^{(|\Theta|)} \}$.
To integrate the masks into the vision encoder, we apply a large negative bias to masked logits. Specifically, for layer $l$ we set 
$\overline{QK}_{(a,b)}^{(l)} = \tfrac{Q_a^{(l)} \cdot K_b^{(l)}}{\sqrt{D}}$ if the attention mask $\mathbf{A}_{(a,b)}^{(r)}=1$, and 
$\overline{QK}_{(a,b)}^{(l)}=-\infty$ otherwise, 
where $Q_a^{(l)}$ and $K_b^{(l)}$ are the query and key vectors of tokens $a$ and $b$, and $D$ is the embedding dimension. 
This ensures that tokens only attend to others within the same SAM-derived region.

\subsection{Model composition based on PVSM}
\label{PVSM}

To improve generalization while remaining training-free, we merge several CLIP variants by averaging their parameters with weights derived from a new metric, the \emph{Prompt Variant Separation Margin (PVSM)}.  Given a set of models $\{\text{Mod}_o\}_{o=1}^O$ with parameters $\phi_o$, we obtain fused parameters $\phi_f = \sum_{o=1}^O w_o \phi_o ,
$ 
where the weights $w_o$  are based on a new metric that measures the similarity of text embeddings obtained for  augmented text prompts for the same class. 
Images are not considered to compute this metric (and, thus, the weights)  because encoding  augmented images across a large dataset would be computationally too expensive.

For each class $c$, we define a base prompt \( pr_c \). 
To introduce lexical variation, we define a set \(\text{Syn}_c = \{ \text{s}_{c,n_s}\}_{n_s = 1}^{N_s}\) of $N_s$ class-specific synonyms $\text{syn}_{c,n_s}$ 
and sets of prefixes \( \Pi = \{ \pi_{n_\pi}\}_{n_\pi = 1}^{N_\pi} \) and 
suffixes \( \Sigma  = \{ \sigma_{n_\sigma}\}_{n_\sigma = 1}^{N_\sigma}\), 
where each prefix \( \pi_{n_\pi} \) is a natural language phrase that precedes the synonym, and each suffix \( \sigma_{n_\sigma} \) is a phrase that follows the synonym.
For every class $c$, we randomly generate \( K \) ($K$ is a hyperparameter)  natural language variants \( v_{c,z} \), $z=1, ..., K$ by combining a random prefix \( \pi_{z_\pi} \in\Pi \), a synonym \( s_{c,z_s} \in \text{Syn}_c \), and a random suffix \( \sigma_{z_\sigma} \in\Sigma\) as \( v_{c,z} = \pi_{z_\pi} + \texttt{" of "} + \text{syn}_{c,z_s} + \texttt{" "} + \sigma_{z_\sigma} \), $z, z_\pi, z_s, z_\sigma = 1, ..., K$.
The resulting variant set of $K$ prompts $v_{c,z}$ for class \( c \) is denoted by \( \mathcal{V}_c = \{ v_{c,1}, \ldots, v_{c,K} \} \).
Given a pre-trained CLIP model \( \text{Mod}_o \), each prompt variant \( v_{c,z} \in \mathcal{V}_c \) is tokenized and encoded as \( \mathbf{t}_{c,z}^o = \frac{\text{Mod}_o(\texttt{tokenize}(v_{c,z}))}{\|\text{Mod}_o(\texttt{tokenize}(v_{c,z}))\|_2} \)\,, where \texttt{tokenize} refers to the CLIP preprocessing of text input for the model’s encoder.
Let \( \mathbf{T}_c^o = \{ \mathbf{t}_{c,1}^o, \ldots, \mathbf{t}_{c,K}^o \} \) be the embeddings for class \( c \) and model \( \text{Mod}_o \). We compute the intra-class similarity \( \mu_{\text{intra}}^{(c),o} \) as the average cosine similarity across all unordered pairs within \( \mathbf{T}_c^o \), i.e., \( \mu_{\text{intra}}^{(c),o} = \frac{2}{K(K-1)} \sum_{1 \leq m' < n' \leq K} \langle \mathbf{t}_{c,m'}^o, \mathbf{t}_{c,n'}^o \rangle \), and the inter-class similarity \( \mu_{\text{inter}}^{(c),o} \) as the average similarity between embeddings of class \( c \) and those of all other classes \( c' \neq c \), i.e., \( \mu_{\text{inter}}^{(c),o} = \frac{1}{K^2(C - 1)} \sum_{c' \neq c} \sum_{i'=1}^K \sum_{j'=1}^K \langle \mathbf{t}_{c,i'}^o, \mathbf{t}_{c',j'}^o \rangle \).
The separation margin for class \( c \), defined as \( \delta^{(c),o} = \mu_{\text{intra}}^{(c),o} - \mu_{\text{inter}}^{(c),o} \), reflects how tightly grouped the class embeddings are and how distinct they are from other classes,  indicating how well the model \( \text{Mod}_o \) has learned the underlying class concepts. 
We define the margin for model \( \text{Mod}_o \) as:
\begin{equation}
\text{PVSM}_o   = \frac{1}{C} \sum_{c =1}^C \delta^{(c),o}  
\end{equation}
and use it to define a weight \(w_o\) of \( \text{Mod}_o \)
as the normalized separation margin, thus \( w_o = \text{PVSM}_o / \sum_{u'=1}^O \text{PVSM}_{u'} \).
The parameters \( \phi_{\text{f}} \) of the fused model \( \text{Mod}_{\text{f}} \),  are obtained via a linear combination of the individual model parameters \(\phi_o\): \( \phi_{\text{f}} = \sum_{o=1}^O w_o \cdot \phi_o \).
This formulation enables the weighted interpolation of models in parameter space, with the weights indicating how well a model is able to produce meaningful text embeddings from varying prompts. 
\section{Experimental setup}
\label{Experiments}

\noindent\textbf{Datasets: }
We evaluate our method on the validation set of three high-resolution RS benchmark datasets:
Potsdam \cite{sumbul2019bigearthnet} consists of orthophotos with a ground sampling distance (GSD) of 5 cm and 6 classes, UDD5 \cite{chen2018large} consists of low-altitude oblique UAV images (4 + 1 classes) and OpenEarthMap ~\cite{xia2023openearthmap} provides 0.25–0.5 m satellite/aerial images (8 + 1 classes) from various regions on earth.
Compared to the datasets used for training CLIP, GeoRSCLIP and RemoteCLIP (we use their model parameters in our experiments), vehicles and roads are far finer in Potsdam, UDD5 diverges from the straight-down and medium-resolution satellite views and OpenEarthMap covers additional land-cover categories and varied sensors.

\vspace{1ex}\noindent\textbf{General Setup:} All experiments employ the CLIP-L/14 backbone. 
We parametrize our model as a weighted ensemble of RemoteCLIP \cite{liu2024remoteclip1} and GeoRSCLIP \cite{zhang2024rs5m}, 
selecting them because they supply pretrained weights that are compatible with our used architecture. 
RemoteCLIP was pretrained on 828,725 image–caption pairs automatically generated from 17 public datasets spanning satellite and UAV platforms with GSDs from 5\,cm to 1\,m \cite{liu2024remoteclip1}. 
GeoRSCLIP was pretrained on RS5M, a dataset of 5 million RS image–text pairs, comprising roughly 3 million web-filtered aerial  images and 2 million captioned satellite and aerial scenes from BigEarthNet, FMoW, and MillionAID \cite{zhang2024rs5m}.
The weights for Remote-CLIP and GeoRSCLIP are set to 0.37 and 0.63, determined according to Section~\ref{PVSM}. 
Input images are divided into tiles of 224 × 224 pixels. As these tiles may not fully capture the spatial context of all objects, we use a sliding window approach with a stride of 50 pixels 
 and average the probabilities of a pixel across all overlapping tiles. 
Our vision encoder has $L=24$ layers in total and we apply attention masking to the final \( |\Theta_r| = 6 \) layers. 
The SAM hyperparameters \( \Theta_r\) controlling the generation of masks at varying levels of granularity are detailed in Appendix A.1. 
All hyperparameters were determined using 5\% of the Potsdam training set. 
Details on the text prompts we have used are given in Appendices A.2-A.4. 


\vspace{1ex}\noindent\textbf{Evaluation strategy: }
We evaluate our method using the mean Intersection over Union (mIoU), and compare it against established training-based OVSS frameworks for RS, i.e., SegEarth-OV \cite{li2024segearth} and the method proposed in \cite{cao2024open}. 
As, to the best of our knowledge, there are no training-free OVSS RS approaches, 
we also compare our approach to training-free methods designed for general-purpose segmentation. 
These baselines are initialized with the original CLIP weights, following prior work \cite{li2024segearth}. 
Additionally, we compare our method against a naive CLIP-based baseline by computing cosine similarity between the image and patch tokens. 
%
We conduct ablation studies utilizing the pretrained CLIP, RemoteCLIP and GeoRSCLIP models, assessing the zero-shot semantic segmentation performance of each model individually and exploring different variants of model composition (cf.\  Section~\ref{PVSM}).

\section{Results and discussion}

\begin{figure*}[t]
    \centering
    \includegraphics[width=0.9\textwidth]{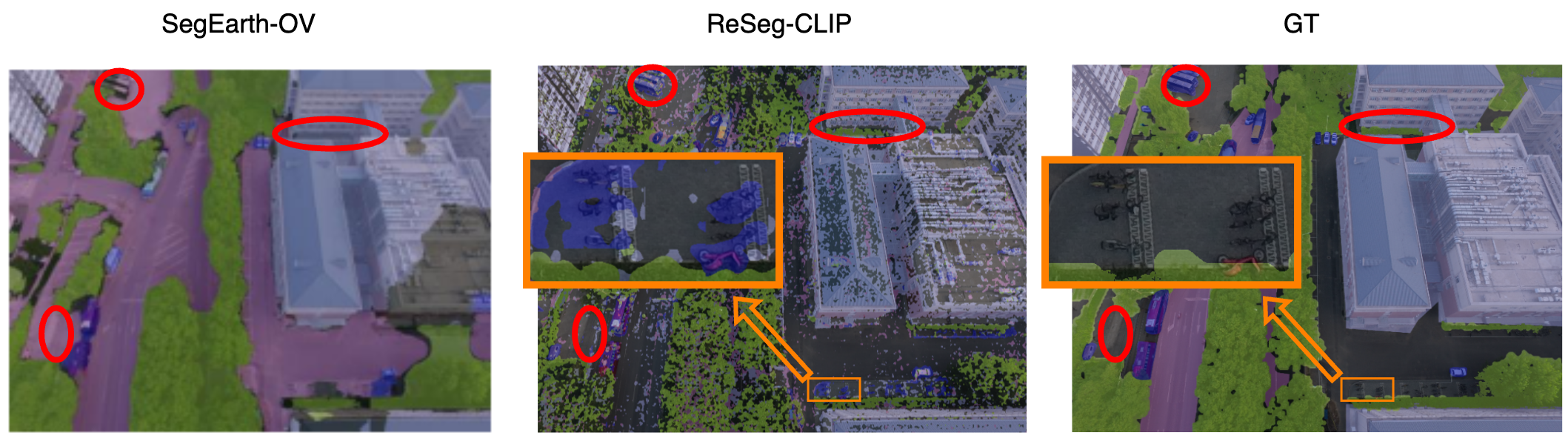}
    \vspace{-0.5em}
   \caption{Results on the UDD5 dataset. SegEarth-OV yields more homogeneous masks, our method offers better class distinction in adjacent regions (red circles). Despite some interpolation-induced noise, our model effectively detects mislabeled areas 
   (orange square).}
    \label{fig:qultiveres}
\end{figure*}

\noindent\textbf{Method performance and comparison: } 
Comparing the results of our method and those of other RS OVSS frameworks (see Tab.~\ref{tab:rs-methods-comparison}), 
our method achieves an 8 percentage points (pp) higher mIOU on the Potsdam dataset compared to \cite{cao2024open}, 
and 7.4\,pp to 8.8\,pp lower mIoU  compared to SegEarth-OV.
This performance gap can be attributed to the use of FeatureUp. This effect is also evident in Figure~\ref{fig:qultiveres}, where the label map generated by SegEarth-OV appears more consistent and homogeneous. 
In contrast, our method achieves more precise spatial localization and clearer class distinction in adjacent regions (red circles), and also demonstrates robustness in scenarios with mislabeled areas (orange square).
While effective, FeatureUp requires to be trained, which conflicts with our training-free design and hinders a fully fair comparison.
However, as FeatureUp is model-agnostic, 
it could be optionally integrated into our method.

\label{sec:results}
\begin{table}[t]
\small
\centering

\setlength{\tabcolsep}{3pt}
\renewcommand{\arraystretch}{1.0}
\begin{tabular}{l|c|c|c|c}
\toprule
\textbf{Method} &
\multicolumn{1}{c|}{\textbf{Potsdam}} &
\multicolumn{1}{c|}{\textbf{UDD5}} &
\multicolumn{1}{c|}{\textbf{OEM}} &
\textbf{Training}
\\
\midrule

Cao et al. \cite{cao2024open} & 30.3 & - & - &  \checkmark \\
Ye et al. \cite{ye2025towards} & 45.7 & - & - &  \checkmark \\
SegEarth-OV & {47.1} & {50.6} & 40.3 &   \maltese \\
\midrule
\textbf{ReSeg-CLIP } & {38.3} & {43.2} & {32.4} &\ding{55}\\
\midrule
\bottomrule
\end{tabular}
\vspace{1ex}
\caption{Comparison of mIoU [\%] across OVSS RS methods. Last column: \(\checkmark\): full-network training; \maltese: training of the upsampling module; \(\times\): no training.}
\label{tab:rs-methods-comparison}
\end{table}

\begin{table}[t]
\centering
\small
\setlength{\tabcolsep}{4pt}
\renewcommand{\arraystretch}{1.2}
\begin{tabular}{l|c|c|c|c|c|c|c}
\toprule
\textbf{Dataset} &
\textbf{CLIP} &
\textbf{MaskCLIP} &
\textbf{SCLIP} &
\textbf{GEM} &
\textbf{ClearCLIP} &
\textbf{C+P} &
\textbf{Ours}
\\
\midrule
\textbf{Potsdam} & 14.5 & 31.7 & 36.6 & 36.5 & \textcolor{red}{40.9} & 18.8 &  \textcolor{blue}{38.3} \\
\textbf{UDD5}    &  9.5 & 32.4 & 38.7 & 41.2 & \textcolor{blue}{41.8} & 15.0 & \textcolor{red}{43.2} \\
\textbf{OEM}     & 12.0 & 25.1 & 29.3 & \textcolor{red}{33.9} & 31.0 & 15.3 & \textcolor{blue}{32.4} \\
\bottomrule
\end{tabular}
\vspace{1ex}
\caption{Comparison of mIoU [\%] (best in red, second best in blue) across training-free general-purpose methods. C+P refers to ReSeg-CLIP without using any SAM-based attention masks in the vision encoder, i.e., $|\Theta|=0$.}
\label{tab:GS-method}
\end{table}

\begin{table*}[t]
\centering
\small

\setlength{\tabcolsep}{5pt}
\renewcommand{\arraystretch}{1.0}
\begin{tabular}{l|c|cccccc|c}
\toprule
\textbf{Parametrization} & \textbf{Weighting} & \textbf{0} & \textbf{1} & \textbf{2} & \textbf{3} & \textbf{4} & \textbf{5} & \textbf{mIoU} \\
\midrule
CLIP \cite{radford2021learning} & - & 25.9 & 44.4 & 34.6 & 41.5 & 6.1 & 3.4 & 24.5 \\
GeoRSCLIP \cite{zhang2024rs5m}  & - & 34.8 & 54.4 & \textcolor{blue}{50.1} & 51.0 & \textcolor{red}{14.7} & \textcolor{red}{5.6} & 33.0 \\
RemoteCLIP \cite{liu2024remoteclip1} & - & 36.5 & \textcolor{blue}{61.4} & 39.1 & 48.1 & 2.8 & 2.5 & 30.4 \\
\cite{zhang2024rs5m} + \cite{liu2024remoteclip1} & PVSM & \textcolor{red}{41.7} & 60.2 & \textcolor{red}{53.3} & \textcolor{red}{59.3} & 11.3 & 3.7 & \textcolor{red}{38.3} \\
\cite{zhang2024rs5m} + \cite{liu2024remoteclip1} & equal & \textcolor{blue}{39.1} & 56.5 & 49.9 & \textcolor{blue}{55.6} & 10.6 & 3.5 & \textcolor{blue}{35.9} \\ 
\cite{zhang2024rs5m} + \cite{liu2024remoteclip1} + \cite{radford2021learning}  & PVSM &29.0 & 48.9 & 34.9 & 44.1 & 8.1 & 2.8 & 28.9 \\
\bottomrule
\end{tabular}
\vspace{1ex}
\caption{Per-class IoU [\%] and mean IoU (mIoU) [\%] (best in red, second best in blue) on the Potsdam dataset. 
All variants with $|\Theta|=6$ layers with SAM-based attention masks in the vision encoder (cf.~Sec.~\ref{Refining}).
Classes: 0: Artificial Surface, 1: Building, 2: Natural Surface, 3: Vegetation, 4: Vehicle, 5: Background.}
\label{tab:ablation-study}
\end{table*}

\begin{table}[t]
\centering
\setlength{\tabcolsep}{8pt}
\begin{tabular}{c|cccccc}
\hline
$\mathbf{|\Theta|}$ & 0 & 1 & 3 & 6 & 12 & 18 \\
\hline
\textbf{mIoU [\%]} & 18.8 & 25.5 & \textcolor{blue}{33.7} & \textcolor{red}{38.3} & 32.1 & 19.4 \\
\hline
\end{tabular}
\vspace{1ex}
\caption{Effect of number of layers in ReSeg-CLIP's vision encoder with SAM-based attention masks $|\Theta|$ on Potsdam dataset (best in red, second best in blue).}
\label{tab:num_attention_mask_miou}
\end{table}

As listed in Table \ref{tab:GS-method}, compared to a naive CLIP-based model parametrized based on \cite{radford2021learning} and to a variant of ReSeg-CLIP without using any SAM-based attention masks in the vision encoder (called C+P), our method achieves significantly higher performances across all datasets. These gains can be attributed to our proposed modifications; the refinement of attention maps and the focus on semantically relevant regions. Additionally, our approach outperforms other training-free methods, including MaskCLIP and SCLIP, across all three benchmarks. When compared to GEM, our model performs better on Potsdam and UDD5, with gains of 1.8\,pp and 2.0\,pp, respectively, though it lags on OEM by 1.5\,pp. Relative to ClearCLIP, our method shows superior performance on UDD5 and OEM by 1.4\,pp but falls short by 2.6\,pp on Potsdam. This variation suggests that OVSS models, originally developed for natural image domains, can still generalize reasonably well to RS tasks, though their performance tends to be inconsistent across datasets. In contrast, our method demonstrates greater consistency, achieving best or second-best performance on all benchmarks. This underscores the effectiveness of our SAM-based hierarchical attention mechanism compared to modifications of the attention module without semantic guidance for inter-patch attention. 

Analyzing the per-class IoU values given in Table~\ref{tab:ablation-study}, it can be seen that our method achieves good results of about 60\% for the classes \textit{Building} and \textit{Vegetation}. On the other hand, the IoU values for \textit{Vehicle} and \textit{Background} are particularly low, which is also the case for all other methods listed. This indicates that segmenting smaller objects and obtaining a meaningful representation of such a heterogeneous class as background poses particular challenges to training-free methods in general and requires further investigations.

\vspace{1ex}\noindent\textbf{Ablation studies:}
We conducted two ablation studies. The results in Table~\ref{tab:ablation-study} show that initializing our model with the original CLIP model weights performs poorly (24.5\% mIOU),  while using the weights of RemoteCLIP and GeoRSCLIP achieves a 5.9\,pp and 8.5\,pp higher mIOU, respectively. 
As CLIP was not exposed to RS data during training, this is an expected outcome. %
Among all pairwise combinations, merging RemoteCLIP and GeoRSCLIP yields the best results, highlighting the effectiveness of fusing complementary information for more generalizable representations. 
Employing the proposed PVSM strategy increases the mIoU by 2.4\,pp compare to equal weighting, demonstrating the semantic expressiveness of our metric. 
In contrast, combining all three models  performs worse than the best pairwise combination. We hypothesize that this is due to parameter oversmoothing, which may suppress critical neural activations and dilute the individual strengths of the fine-tuned models.
Our second ablation study investigates the impact of varying the number of final layers in the vision encoder that restrict attention based on SAM-derived masks. The results show that increasing this number up to 6 progressively improves the mIoU 
up to 38.3\% (cf.~Tab.\,~\ref{tab:num_attention_mask_miou}). This highlights the effectiveness of hierarchical feature aggregation guided by SAM masks. However, when the number of masked layers is increased beyond 6, the performance drops. This indicates that preserving global context in early layers while applying localized attention in later ones is optimal for the segmentation performance.

\section{Conclusion}
In this work, we introduce ReSeg-CLIP, a fully training-free method for OVSS of RS images. Our method tackles two challenges of VLMs: disrupted patch-level attention in dense prediction tasks and poor generalization across domains. 
To address these challenges, we propose a hierarchical attention masking strategy that applies multi-scale SAM-generated masks at different vision encoder depths 
and a weight-space model composition technique 
using weights derived from our novel data-driven PVSM metric. 
Extensive experiments across three RS benchmarks demonstrate improvements in accuracy and robustness due to our hierarchical masking strategy and model composition technique, respectively. Combining both contributions, our method achieves promising results particularly for buildings and vegetation, outperforms existing training-free approaches and achieves competitive results with partially trained ones. 
Future work following the training-free paradigm may explore incorporating image-aware model fusion criteria, optimizing hierarchical masking for efficiency, and improving the alignment of masks with the true semantic boundaries.

\bibliography{egbib}
\end{document}